\documentclass[submission,copyright,creativecommons]{eptcs}

\usepackage{iftex}

\ifpdf
  \usepackage{underscore}         
  \usepackage[T1]{fontenc}        
\else
  \usepackage{breakurl}           
\fi

\usepackage{libertine} 
\usepackage{libertinust1math}

\usepackage{times}
\usepackage{graphicx}

\usepackage{amsmath,amssymb,amstext, amsthm, amsfonts}
\usepackage{tikz-cd}
\usetikzlibrary{calc}
\usetikzlibrary{decorations.markings}

\usepackage{url,hyperref}
\usepackage{csquotes}
\usepackage{caption}
\usepackage{subcaption}
\usepackage{paralist}
\usepackage{enumerate}
\usepackage{booktabs}

\DeclareMathAlphabet{\mathcal}{OMS}{cmsy}{m}{n}

\newcommand{\code}[1]{\texttt{#1}}

\newtheorem*{proposition*}{Proposition}
\newtheorem{example}{Example}[section]

\DeclareMathOperator{\ob}{ob}

\DeclareMathOperator{\simi}{sim}
\DeclareMathOperator{\braid}{orth}

\newcommand{\cat}[1]{\mathcal{#1}}
\newcommand{\catname}[1]{{\normalfont\textbf{#1}}}
\newcommand{\Set}{\catname{Set}}

\newcommand{\VSA}{\catname{VSA}}

\newcommand{\R}{\mathbb{R}}
\newcommand{\C}{\mathbb{C}}
\newcommand{\IMon}{\textcolor{red}{\otimes}}

\title{Developing a Foundation of Vector Symbolic Architectures Using Category Theory}
\author{
Nolan Peter Shaw
\institute{Cheriton School of Computer Science\\
University of Waterloo}
\email{nolan.shaw@uwaterloo.ca}
\and
P. Michael Furlong
\institute{Centre for Theoretical Neuroscience\\
Systems Design Engineering\\
University of Waterloo}
\email{michael.furlong@uwaterloo.ca}
\and
Britt Anderson
\institute{Department of Psychology\\
University of Waterloo}
\email{britt@uwaterloo.ca}
\and
Jeff Orchard
\institute{Cheriton School of Computer Science\\
University of Waterloo}
\email{jorchard@uwaterloo.ca}
}
\def\titlerunning{Category theory for VSAs}
\def\authorrunning{Nolan Peter Shaw, P. Michael Furlong, Britt Anderson, Jeff Orchard}
\begin{document}
\maketitle

\begin{abstract}
Connectionist approaches to machine learning, \emph{i.e.} neural networks, are enjoying a considerable vogue right now. However, these methods require large volumes of data and produce models that are uninterpretable to humans. An alternative framework that is compatible with neural networks and gradient-based learning, but explicitly models compositionality, is Vector Symbolic Architectures (VSAs). VSAs are a family of algebras on high-dimensional vector representations. They arose in cognitive science from the need to unify neural processing and the kind of symbolic reasoning that humans perform. While machine learning methods have benefited from category-theoretical analyses, VSAs have not yet received similar treatment. In this paper, we present a first attempt at applying category theory to VSAs. Specifically, We generalise from vectors to co-presheaves, and describe VSA operations as the right Kan extensions of the external tensor product. This formalisation involves a proof that the right Kan extension in such cases can be expressed as simple, element-wise operations. We validate our formalisation with worked examples that connect to current VSA implementations, while suggesting new possible designs for VSAs.
\end{abstract}

\section{Introduction}
\label{sec:intro}
The rapid adoption of machine learning in software engineering has largely outpaced the rate at which we understand many learning algorithms. To better understand these algorithms, two main approaches have been taken. The first is to make complex models more \enquote{interpretable} to human users. The second, more recent approach, has been to improve our understanding of the mathematical foundations of machine learning. While the mechanics of differentiation and matrix multiplication have been well-understood for a long time, these provide little perspective on the \emph{emergent} behaviour and structure of large-scale algorithms.

In the past few years, considerable strides have been made in developing a category-theoretic perspective on machine learning and deep learning~\cite{cruttwell2022categorical, gavranović2024fundamental, shiebler2021category}. A lot of this work has focused on supervised learning, where models learn through exposure to input/output training pairs. Here, lenses have been identified as providing the right structure for explaining the bi-directional flow of forward-propagating computations and backward-propagating errors. Parameterising these lenses yields a means to modify the computation of a model.

Less progress has been made on understanding learning (and, more generally, cognition) from the perspective of information storage and association. Artificial feed-forward neural networks do a good job of capturing the hierarchical structures in brains, but brains are very complicated, and neural connectivity is often highly recurrent and seemingly chaotic. This complexity has prompted some researchers to look at neuronal computation at the level of \enquote{populations} of neurons and the dynamics contained within. Examples include reservoir computing~\cite{schrauwen2007overview, tanaka2019recent}, neural engineering frameworks~\cite{eliasmith2013build}, and the focus of this paper: vector symbolic architectures, or \enquote{VSAs}.

As the name implies, VSAs are fundamentally vector spaces that aim to encode symbols with an associated \enquote{meaning} as vectors. They possess a means of comparing how similar vectors are, and two operations---binding/unbinding and bundling---that produce new vectors corresponding to more sophisticated symbols. Maintaining (near-)orthogonality is critical to ensuring that symbols can still be distinguished from one another. The advantages of VSAs are evident in their ability to store information: given a vector space of dimension $d$, $2^d$ near-orthogonal vectors can be encoded with a high degree of noise-resistance~\cite{hecht1994context}. Further, for researchers who want to study learning as it occurs in biological brains, VSAs have been implemented in spiking neuron models\cite{eliasmith2012large,eliasmith2013build,dumont2023exploiting,furlong2023modelling,orchard2023hyperdimensional,gosmann2016optimizing,komer2019neural,stewart2010symbolic,Dumont2019,renner2022sparse,frady2019robust,Frady2022,frady2021variable}, demonstrating their biological plausibility.

VSAs also aim to describe cognitive features such as \emph{compositionality} that are absent from some other connectionist models~\cite{fodor1988connectionism}. In spite of this ambition, discussion of VSAs in categorical terms is largely absent from the literature; we provide a literature review in Appendix~\ref{app:lit-rev}. This paper is an initial step in understanding VSAs from a category-theoretic perspective. We provide a list of desiderata for properties of VSAs, similar to the work done in~\cite{cruttwell2022categorical} for supervised learning. Our primary contribution is generalising from vectors to co-presheaves, and then deriving \enquote{optimal} VSA operations using Kan extensions. This formalisation determines the proper choice of bind and bundle operations.

\section{Background}
\label{sec:background}
VSAs\footnote{Terminology attributed to~\cite{Gayler2003}, VSAs are also known as Hyperdimensional Computing (HDC)\cite{kanerva2009hyperdimensional}, and more recently, Vector Symbolic Algebras\cite{furlong2023bridging}.} address the problem of how human behaviour can be characterised by rules and symbolic reasoning, but be implemented in neural networks, which rely on manipulating vector representations and engage in similarity-based reasoning~\cite{smolensky1988proper}. VSAs use randomly generated vectors to represent variables and the values they take on (also called \enquote{slots} and \enquote{fillers}). These base symbols are composed into higher-level representations using a fixed set of algebraic operations.

\subsection{VSA Definition}
\label{sec:vsa-def}
Individual architectures can be described by the types of vectors that they operate on and corresponding operations: similarity, bundling, binding, unbinding, and braiding\cite{gayler1998multiplicative}. Frady \emph{et al.} note that bundling and binding \enquote{are dyadic operations that form a ring like structure}~\cite{frady2021variable}. Additionally, there are operations which may assist in implementing the algebra, namely the (pseudo-)inverse of vectors, and \enquote{cleanup} operations for de-noising. While not commonly emphasised in VSA literature, vectors can be multiplied by scalars, including negative numbers.

For any particular VSA, the vector space, or domain manifold within, determines the choice of operators and the space that base vectors are generated in, which can include $\{-1,1\}^{d}$\cite{gayler1998multiplicative}, $\{0,1\}^{d}$\cite{kanerva1996binary,rachkovskij2001binding,Laiho2015,frady2021variable}, $\mathbb{R}^{d}$\cite{plate1995holographic,gallant2013representing,gosmann2019vector}, and $\mathbb{C}^{d}$\cite{plate1994distributed,frady2019robust,frady2021variable}, where $d$ represents the dimensionality of our space, sometimes in the order of $10, 000$. It is also not unusual for there to be further constraints placed on base vectors. For example, the Fourier Holographic Reduced Representations (FHRR; \cite{plate1995holographic}) algebra (see Table~\ref{tab:vsa-names} for a list of VSA names and abbreviations) requires that the complex numbers comprising the vector have magnitude one, forcing all datapoints to live on a hypertorus, and HRR vectors are normalized to unit length. The Binary Spatter Code (BSC; \cite{kanerva1996binary}) and Multiply-Add-Permute (MAP) VSAs require atomic vectors that live on the vertices of hypercubes. Other VSAs enforce sparsity~\cite{rachkovskij2001binding} or a specific block structure~\cite{Laiho2015,frady2021variable} in their atomic vectors.  For the purposes of this document we will denote the space of vectors as $X$.

\begin{table}[ht]
    \centering
    \begin{tabular}{l|ll}
         \textbf{Initialism} & \textbf{VSA Name} & \textbf{Citation(s)}  \\
         \toprule
         TPR & Tensor Product Representations &~\cite{smolensky1990tensor}\\
         \midrule
         MAP & Multiply Add Permute &~\cite{gayler1998multiplicative} \\
         HRR & Holographic Reduced Representations &~\cite{plate1994distributed,plate2003holographic} \\
         FHRR & Fourier Holographic Reduced Representations & " \\
         BSC & Binary Spatter Code &~\cite{kanerva1996binary} \\
         BSDC & Binary Sparse Distributed Codes (or Representations) &~\cite{rachkovskij2001representation,rachkovskij2001binding,Laiho2015,frady2021variable}\\
         VTB & Vector-derived Tensor Binding &~\cite{gosmann2019vector}\\
         MBAT & Matrix Binding of Additive Terms &~\cite{gallant2013representing,tissera2014enabling}\\
         \bottomrule
    \end{tabular}
    \caption{A list of the considered VSAs, clarifying their naming conventions, and providing references to important early papers.}
    \label{tab:vsa-names}
\end{table}

\subsection{VSA Operations}

\paragraph{Similarity:} $\simi : X \times X \rightarrow \mathbb{R}$. This function takes two vectors and returns a number that measures similarity. Quite commonly, VSA similarity is provided by cosine similarity. In the case of binary or bipolar vectors Hamming distance or overlap are used. In all cases, similarity is bounded, with extremum (usually one) indicating equivalence between vectors. While we may not always be working with exact values, for the purposes of this document we will denote equivalence between vectors as $\simi(x, y) = 1$, and indicate maximum dissimilarity between vectors as $\simi(x, y) = 0$. We use $x \sim y$ to denote two vectors that are similar.

\paragraph{Bundling:} $\oplus : X \times X \rightarrow X$. The bundling operation takes two vectors and produces a new vector that, importantly, maintains some degree of similarity to its constituent parts. Given a bundle $z = x \oplus y$, we expect that both $\simi(x, z)$ and $\simi(y, z)$ to be high, although possibly not as high as $\simi(x, x)$ or $\simi(y, y)$. We note that for some VSAs, the bundling operation comes with some additional normalisation term which ensures that the magnitude of the bundles never exceed some specified range (\emph{e.g.}, the unit hypersphere, vertices of the unit hypercube).

\paragraph{Binding:} $\otimes: X\times X \rightarrow X$. Binding transforms two vectors into a new vector that is not similar to either constituent element. Given a bound vector $z = x\otimes y$, we expect that $\simi(x, z) \sim 0 \sim \simi(y, z)$. While it is commonly the case, binding is not always commutative or associative in all VSAs (\emph{ex.} Smolensky's tensor product~\cite{plate1995holographic,gayler1998multiplicative,frady2021variable}).

\paragraph{Unbinding:} $\oslash: X\times X \rightarrow X$. The purpose of unbinding is to undo the binding operation, \emph{i.e.}, if $z = x \otimes y$, then $x \oslash z = y$ and $z \oslash y = x$. Despite some VSAs distinguishing this operation from binding, what we really require is to have some value $k = x^{-1}$ such that $k \otimes z = y$ (or some right inverse for the other case). Depending on the chosen algebra and vector space, the vectors can be \emph{self-inverse} or not, with respect to binding~\cite{schlegel2022comparison}. In practice, \enquote{lossy} binding may mean that only a pseudo-inverse can be found. In this case, then $x^{-1} \otimes z \sim y$, whereas for exact inverses $x^{-1} \otimes z = y$ holds. This problem is often addressed using an additional \enquote{cleanup} operation. We expand on the relationship between binding, unbinding, and inverses in Section~\ref{sec:act}.

\paragraph{Braiding:} $\braid : X \rightarrow X$. Braiding (alternatively, \emph{hiding}) is an invertible unitary operation that produces a new vector that is dissimilar to the input vector, \emph{i.e.}, $\simi(x, \braid(x)) \sim 0$. Braiding was introduced by~\cite{gayler1998multiplicative} to deal with a number of problems in VSAs that only supplied binding and bundling. Specifically, if the binding operation is commutative, or if a given vector is its own inverse, then it is not possible to impose hierarchical structure on encoded representations. As we will discuss below, the need for a braiding operation may be an artefact of choosing a commutative binding operator.

\subsection{Survey of Commonly Used VSAs}
\label{sec:vsa-overview}
Table~\ref{tab:vsa-names} provides a list of VSA acronyms, and references that describe them. Smolensky's Tensor Product Representations\cite{smolensky1990tensor} is the first VSA. It uses the tensor product as a binding operation that is not commutative, but is self-inverse. However, the tensor product has the practical limitation that the dimensionality of the vector representation grows exponentially with the number of applications of the binding operator. Consequently, when Gayler~\cite{Gayler2003} defined vector symbolic architectures, he focused on algebras with dimensionality-preserving operations. Schlegel \emph{et al.}\cite{schlegel2022comparison} surveyed different VSAs, comparing different operations, which we produce a modified version of in Table~\ref{tab:comparison}.

\begin{table}[t]
    \centering
    \makebox[\textwidth][c]{
    \begin{tabular}{l|p{0.15\textwidth}p{0.075\textwidth}p{0.11\textwidth}p{0.15\textwidth}p{0.12\textwidth}p{0.12\textwidth}p{0.14\textwidth}}
         \textbf{Algebra} & \textbf{Base Vector} & \textbf{Domain} & \textbf{Similarity} & \textbf{Bundling} & \textbf{Binding} & \textbf{Inverse} & \textbf{Braiding}\\
         \toprule
         TPR & $\{-1,1\}^{d}$ &$\mathbb{R}^{d}$ & cosine similarity & Vector addition & Tensor product & $x^{-1} = x$ & -- \\
         \midrule 
         MAP-I & $\{-1,1\}^{d}$ & $\mathbb{Z}^{d}$ & " & Elem. addition& Hadamard prod. & $x^{-1} = x$ & permutation \\
         MAP-B & " & $\{-1,1\}^{d}$ & " & " + threshold & " & " & " \\
         MAP-C & $[-1,1]^{d}$ & $\mathbb{R}^{d}$ & " & " + cutting & " & " & " \\
         \midrule
         FHRR & $\mathrm{e}^{i\mathbf{a}}$ $(a_{i})~\in~[-\pi,\pi]$& $\mathbb{C}^{d}$ & inner product & " & " & $(x^{-1})_{i}=1/x_{i}$ & permutation or bind with random vector\\
         HRR & $(x_{i}) \sim \mathcal{N}(0,\frac{1}{d})$ & $\mathbb{R}^{d}$ & cosine similarity & " + normalization & circular convolution & $(x^{-1})_{i}=$ $x_{(d-i+1)\mod d}$& " \\
         \midrule
         MBAT & " & $\mathbb{R}^{d}$ & " & " + normalization & Matrix mul. & Matrix Inv. & --\\
         VTB & " & $\mathbb{R}^{d}$ & " & " + normalization& VTB & transpose VTB & --\\
         \midrule
         BSC & $\{0,1\}^{d}$& $\{0,1\}^{d}$ & Hamming distance & AND & XOR & $x^{-1}=x$ & permutation\\
         BSDC-S & " & " & Normalized Overlap & OR & Shifting & Reverse shift & -- \\
         BSDC-SEG & " & " & " & " & Segment Shift & " & -- \\
         BSDC-CDT & " & " & " & " & OR + CDT & -- & -- \\
         \bottomrule
    \end{tabular}
    }
    \caption{We provide a modified version of Table 1 from Schlegel \emph{et al.}~\cite{schlegel2022comparison}. We adopt the notation of Schlegel \emph{et al.}, annotating the MAP VSAs to indicate the different domains vectors are permitted to exist in. Note that these different spaces come equipped with slightly different bundling operations, to ensure the produced vectors stay in the desired space. Similarly, we use Schlegel \emph{et al.}'s notation to distinguish BSDC VSAs that use different binding operations.}
    \label{tab:comparison}
\end{table}

Note in Table~\ref{tab:comparison} that not all VSAs require a braiding operator. Typically, braiding is required when the binding operation is commutative. Without being able to distinguish between $x \otimes y$ and $y \otimes x$ using the similarity operator, it becomes difficult to implement more complex structures like trees or graphs, as we will discuss below. This is further compounded if the binding operation admits a self-inverse. In the case of the (F)HRR, VTB, and MBAT algebras, one way to impose some structure is to repeatedly bind a vector with itself; in the case of self-inverting representations this would not be possible. 

The binding operator for the TPR, MAP, (F)HRR, BSC, and MBAT VSAs are fairly straightforward, but it is worth explaining binding for the VTB and the BSDC VSAs. For VTB, when two vectors, $x,y$, are bound, one vector is used to construct a matrix, $V(x)$, with a block diagonal structure, and then the bound vector is constructed $z=V(x)y$. For BSDC-S vectors are bound by circularly shifting one vector by an integer number of steps derived from the other vector. The BSDC-CDT (Context-Dependent Thinning)\cite{rachkovskij2001representation,rachkovskij2001binding} binding operator is unique in that it produces new vectors that are not orthogonal to their inputs. For the BSDC-SEG VSA, the vectors are divided up into blocks, and the shifting process is applied block-wise\cite{Laiho2015}, which is equivalent to circular convolution applied block-wise for maximally sparse blocks\cite{frady2021variable}. Frady \emph{et al.}~\cite{frady2021variable} proposed other sparsity-preserving binding operations for sparse block codes.

\subsection{Example Problem: Composing Functions}
There are a number of ways to compute functions in a VSA. One way is to take the set definition of a function, $F = \left\{(x,f(x))\mid x \in X\right\}$ and encode it as a bundle over the domain and range of the function:
\begin{align}
    F &= \underset{x \in X}{\oplus} \left(x \otimes f(x)\right),\\
    \Leftrightarrow f(x) &\sim  x^{-1} \otimes F.
\end{align}
For computer scientists, this equivalence is really just the idea that functions can be curried and uncurried. To compose two functions, $f$ and $g$, we would write:
\begin{align}
    (f;g)(x) \sim ((x^{-1} \otimes F)^{-1} \otimes G).
\end{align}
In this case, the function value should be approximately equivalent to the value we would have obtained had we just implemented $(f;g)(x)$ directly.

Because some unbinding operations are only approximate and we're manipulating (potentially) large bundles of only pseudo-orthogonal vectors, there is always a risk of cross-talk noise. Consequently, it is sometimes the case that clean-up operations are interjected between stages, \emph{i.e.},
\begin{align}
    (f;g)(x) = \mathrm{cleanup}(\mathrm{cleanup}(x^{-1} \otimes F)^{-1} \otimes G).
\end{align}
Here cleanup projects a noisy vector back to the closest point in the space, $\left\{f \mid f \in F\right\}$. More examples of VSA computations can be found in Appendix~\ref{app:ex}.

\section{Desiderata}
\label{sec:desiderata}
Here we characterise the fundamental properties that enable VSAs to represent concepts and perform computations, and which we require any categorical formalisation to respect. The first two are strictly necessary. The third arises from practical considerations, but any formalisation based on category theory should also capture the cases where it is needed.

\begin{enumerate}[\bfseries 1.]
    \item{\textbf{Similarity:}} The first necessary property of a VSA is the notion of how \enquote{similar} two vectors are (or aren't). We wish to quantify how closely related concepts are to one another. We also need a means of ensuring that a \enquote{collision} hasn't occurred---that is that two vectors aren't so similar that they are difficult to distinguish.
    
    \item{\textbf{Two binary operations:}} We'd like to have an operation that produces a vector that is similar to its operands, and one that creates a vector that is dissimilar. We've been referring to these operations as bundling and binding, respectively, and denote them as $\oplus$ and $\otimes$. Further, we require that it is possible to \enquote{undo} both operations.

    \item{\textbf{Dimension Preservation:}} An important practical concern when building VSAs is that the dimensions of our vectors remain constant. It is understood that two vectors can be bound and bundled without losing any information in Smolensky's TPR however, as mentioned in Section~\ref{sec:vsa-overview}, this results in $d^2$-dimensional vector given $d$-dimensional operands. To alleviate space constraints, practitioners need bind and bundle operations that always produce vectors with a fixed size.
\end{enumerate}

\section{Formalising VSAs Using Category Theory}
\label{sec:act}
VSA require two binary operations. One must be reversible, and the other must distribute over the first. This requires our objects to be at least a ring. Because we want to perform unbinding and unbundling, this ring must be a \emph{division} ring. Note that this formalisation necessitates that binding, and bundling, are invertible. The tendency of VSA architects to specify separate and distinct operations for unbinding and unbundling is a form of conceptual obfuscation.

Our strategy is to decouple our vectors into indices and values. This generalises from vectors to co-presheaves\footnote{\enquote{Co-presheaves} are conceptual identical to covariant functors. We prefer the term co-presheaves to highlight that we're ultimately concerned with the objects in $[\cat I, \cat V]$.}. With this, we can isolate the division ring structure in our VSAs as arising from the division ring structure of the category from which we draw our values. This decoupling gives us the capacity to include both dimension preserving and non-dimension preserving VSAs. We formalise the bind and bundle operations as right Kan extensions to the external tensor product and sum, respectively.

Finally, if we impose that our co-presheaves exist as a dagger category, and that the category of values is not just a division ring, but also enriched, then we recover multiple implementations of the various similarity measures used in practice. 

\subsection{Generalising to Co-presheaves}
First, we generalise from vectors to functors $v \colon \cat I \to \cat V$ from a (closed) Cartesian monoidal category, $(\cat I, \IMon, I)$, to a division ring category, $(\cat V, \cdot, +, 1, 0)$. The action of $v$ on an object $i \colon \cat I$ is written as $v_i$. Since the fundamental \enquote{vectors} of a VSA are our primary focus, we'll refer to these functors as co-presheaves to highlight our interest in the arrow category, $[\cat I, \cat V]$.

Intuitively, $\cat I$ can be understood as an \emph{indexing} category, while $\cat V$ is the category from which we draw our values. In the case of traditional, real-valued vectors in three dimensions, $\cat I$ would be the discrete category whose objects are the natural numbers, $1$, $2$, $3$, while $\cat V$ would be $\R$. Hence, a $3$-dimensional vector could be written as $v = [v_0, v_1, v_2]$, reconciling conventional vector representations with our generalisation.

\subsection{External Tensor Product and Sum}
In the remaining work, we will restrict our discussion to the bind operation, $\otimes$, since all the following arguments apply equivalently to bundle, $\oplus$. With our new generalisation to co-presheaves, remember that our goal is to have two operations (bind and bundle) of type  
\[\otimes, \oplus \colon [\cat I, \cat V] \times [\cat I, \cat V] \to [\cat I, \cat V]. \]

We can get part way there using the external tensor product. The external tensor product of two co-presheaves has type:
\[ \overline{\otimes} \colon [\cat I, \cat V] \times [\cat I, \cat V] \to [\cat I \times \cat I, \cat V] \] for operands, $v, w \colon [\cat I, \cat V]$. Its action on objects, $i, j \colon \cat I \times \cat I$, is given by:
\[ (v \overline{\otimes} w)\langle i, j \rangle = v_i \cdot w_j, \] where $(\cdot)$ comes from the ring operation of $\cat V$. The exact same types and notation apply to the external sum $\overline{\oplus}$ using $(+)$.

In fact, if we don't need bind and bundle to preserve dimensionality (as is the case for Smolensky's TPA), then the external tensor product/sum are respectively the exact operations we need. However, we can do better yet.

\subsection{Kan Extensions}
Our last task is to find a mapping $\cat I \times \cat I \to \cat I$. Applying this map inside $[\cat I \times \cat I, \cat V]$ would give us the desired output of type $[\cat I, \cat V]$. For this, remember that we've specified that $\cat I$ must be monoidal. Thus, we can use $\IMon$ from $\cat I$ to induce the following diagram:
\[
 \begin{tikzcd} [row sep=1cm, column sep=1cm]
 \cat I \times \cat I
 \arrow[rr, "v \overline{\otimes} w"]
 \arrow[d, bend right, "\IMon"']
 && \cat V
 \\
 \cat I
  \arrow[urr, bend right=10, "v \otimes w" {yshift=3pt, above}]
 \end{tikzcd}
\]

However, there are many possible ways that $v \otimes w$ could map from $\cat I$ to $\cat V$. What we really want is some \enquote{optimal} choice that preserves as much information as possible from $v \overline{\otimes} w$. Questions of optimisation are often expressed categorically using (co-)limits, and the triangular diagram above prompts us to express this limit as a Kan extension. While the left Kan extension would give us the \enquote{freest,} or most \enquote{liberal,} version of $\otimes$ w.r.t. $\overline{\otimes}$, what we really want is the version of $\otimes$ that is as \emph{similar} to $\overline{\otimes}$ as possible. Such a notion is captured by the right Kan (Ran) extension, which can be intuitively understood as the most \enquote{cautious,} or \enquote{conservative,} version of $\overline{\otimes}$ when extended along $\IMon$.

For review: let $\cat C, \cat D$, and $\cat E$ be categories. The right Kan extension of a functor, $F \colon \cat C \to \cat E$, when extended along a functor $e \colon \cat C \to \cat D$ is the (unique up to isomorphism) functor $\text{Ran}_e F \colon \cat D \to \cat E$ along with a natural transformation, $\eta \colon e;(\text{Ran}_e F) \Rightarrow F$. This natural transformation comes with the guarantee that any other $u \colon \cat D \to \cat E$  with its own natural transformation of type $(e;u \Rightarrow F)$ has a unique mapping to $\text{Ran}_e F$. This relationship is captured by the following diagram:
\[
 \begin{tikzcd} [row sep=1cm, column sep=1cm]
 |[alias=C]| \cat C
   \arrow[rr, "F", "" {name=ID, below}]
   \arrow[d, bend right, "e"']
 && \cat E
 \\
 |[alias=D]| \cat D
   \arrow[urr, bend right=45, "u" {xshift=3pt, below}]
   \arrow[urr, bend right=45, "" {name=U, below}]
   \arrow[urr, bend right=10, "\text{Ran}_e F" {yshift=3pt, above}]
   \arrow[urr, bend right=10, "" {name=R, above}]
   \arrow[Rightarrow, "!", shorten=2pt, from=U, to=R]
   \arrow[Rightarrow, bend left, "\eta" {xshift=-8pt, yshift=-3pt, above}, from=D, to=ID]
 \end{tikzcd}
\]

The right (and left) Kan extensions do not always exist for any pair of functors, $F$ and $e$. Though as luck would have it, Kan extensions for the external tensor product do exist! An expression for this extension is given in~\cite{kelly1982basic} as \[(\text{Ran}_e v \overline{ \otimes } w)_i = \int_{j k} \cat I (i, e(j, k)) \pitchfork (v_j \cdot w_k),\] where $A \pitchfork B$ is the power operator and is simply the set of maps between $A$ and $B$. When $e$ is $\IMon$ we have
\begin{equation}
\label{eq:ran-tens}
    (\text{Ran}_{\IMon} v \overline{ \otimes } w)_i = \int_{j k} \cat I (i, j \IMon k)) \pitchfork (v_j \cdot w_k).
\end{equation}
Merging the two previous diagrams gives us
\[
 \begin{tikzcd} [row sep=1cm, column sep=1cm]
 |[alias=II]| \cat I \times \cat I
 \arrow[rr, "v \overline{\otimes} w", "" {name=ID, below} ]
 \arrow[d, bend right, "\IMon"']
 && \cat V
 \\
 |[alias=I]| \cat I
   \arrow[urr, bend right=45, "u" {xshift=3pt, below}]
   \arrow[urr, bend right=45, "" {name=U, below}]
   \arrow[urr, bend right=10, "\text{Ran}_{\IMon} v \overline{\otimes} w" {yshift=5pt, xshift=-1pt, above}]
   \arrow[urr, bend right=10, "" {name=R, above}]
   \arrow[Rightarrow, "!", shorten=2pt, from=U, to=R]
   \arrow[Rightarrow, bend left, "\eta" {xshift=-8pt, yshift=-3pt, above}, from=I, to=ID]
 \end{tikzcd}
\]
and indicates that $v \otimes w := \text{Ran}_{\IMon} v \overline{\otimes} w$ is the bind operation we're looking for. To this end, let's evaluate the mapping into $\text{Ran}_{\IMon} v \overline{\otimes} w$ from an arbitrary $u \colon \cat I \to \cat V$ (we use component-wise notation for clarity):
\begin{align*}
    & \int_i \cat V \big( u_i, (\text{Ran}_{\IMon} v \overline{ \otimes } w)_i\big) \\
    = & \int_i \cat V \big( u_i, \int_{j k} \cat I (i, j \IMon k) \pitchfork (v_j \cdot w_k)\big), & \text{from~\ref{eq:ran-tens}} \\
    = & \int_{i j k} \cat V \big(u_i, \cat I (i, j \IMon k) \pitchfork (v_j \cdot w_k)\big), & \text{by continuity} \\
    = & \int_{i j k} \Set \big( \cat I (i, j \IMon k), \cat V (u_i, v_j \cdot w_k) \big), & \text{by definition of power operator} \\
    = & \int_{i j k} \Set \big( \cat I (i, j) \times \cat I (i, k), \cat V (u_i, v_j \cdot w_k) \big), & \text{since $\IMon$ is Cartesian} \\
    = & \int_{i j k} \Set \big( \cat I (i, j),  \Set (\cat I (i, k), \cat V (u_i, v_j \cdot w_k ) \big), & \text{by currying} \\
    = & \int_{i k}  \Set (\cat I (i, k), \cat V (u_i, v_i \cdot w_k ) \big), & \text{by Yoneda} \\
    = & \int_i \cat V (u_i, v_i \cdot w_i ), & \text{by Yoneda again} \\
    \Rightarrow & (\text{Ran}_{\IMon} v \overline{ \otimes } w)_i \cong v_i \cdot w_i.
\end{align*}
Thus, we have
\begin{equation}
    v \otimes w =\text{Ran}_{\IMon} v \overline{ \otimes } w = \int_i v_i \cdot w_i
\end{equation}
and
\begin{equation}
    v \oplus w =\text{Ran}_{\IMon} v \overline{ \oplus } w = \int_i v_i + w_i.
\end{equation}

Some exciting observations immediately follow from this derivation. First, it shows that the division ring structure of VSAs is inherited from the category of values, $\cat V$, in our co-presheaves. Second, decoupling indices from values means that we can isolate the matter of \enquote{compression} to the mapping of $\cat I \times \cat I \to \cat I$ along $\IMon$ when dealing with co-presheaves on a fixed, finite domain. Finally, this result validates that the widespread practice of using element-wise operations in VSAs is optimal. In fact, many of the alterations to element-wise multiplication/division (i.e. normalisation, thresholding, etc.) are imposed to model certain behavioural effects, such as neuronal saturation, rather than endowing computational advantages.

\subsection{Similarity}
Referring once again to Table~\ref{tab:comparison}, we notice that there's a trend of using the inner product to compute similarity. We can address this by insisting that our co-presheaves, $[\cat I, \cat V]$, possess a \enquote{dagger,} $\dagger$, in $\cat V$. Explicitly, we have \[ \simi(v, w) = \int^i v^{\dagger}_i \cdot w_i. \]

This expression is a co-end that can be further simplified using the $+$ operation of our division ring. In the case of FHRR, the inner product is used directly. Cosine similarity, which is another popular similarity function, is really just a normalised version of the inner product and is captured by this expression as well (same for the Hamming distance used in BSC). Furthermore, by taking $\cat V$ as a $\cat V$-enriched category (for example, with $\leq$ in $\R$), we have a clear way to quantify \emph{how} similar two co-presheaves are with the above co-end.

\subsection{Worked Examples}
\label{sec:examples}
We first verify that using the canonical tensor product in our indexing category recovers the tensor product as our optimal bind for traditional vectors.
\begin{example}
Let $\cat I$ be the discrete category whose objects are natural numbers, with $\IMon$ being the tensor product. Also, let $\cat V = (\R, \cdot, +, 1, 0)$. Then $[\cat I, \cat V]$ are real-valued vectors, with optimal bind and bundle being element-wise multiplication and addition, respectively, within the resulting $d^2$-dimensional tensor of the original $d$-dimensional $v$ and $w$ vectors, as in TPR models.
\end{example}

While this is great, we still haven't ensured that this formalisation holds in cases where we're restricted to a fixed, finite dimension. Without this assurance, our categorical model is infeasible from a data storage perspective. Hence, we need to consider instances where $\cat I$ has finitely many objects. Fortunately, the only requirement for $\IMon$ in our derivation is that it is Cartesian, and any finite poset with a corresponding \enquote{meet} as its $\IMon$ has this property. Consider the following:
\begin{example}
\label{ex:pow-set}
Let $\cat I$ be the category representing the powerset of three elements, with morphisms being set inclusion. It's objects are the eight subsets of the underlying set and $\IMon$ is set intersection. Then the mapping $\IMon \colon \cat I \times \cat I \to \cat I$ produces the original eight-object lattice structure of the powerset.
\end{example}
Note that in this example there is no mention of $\cat V$. This is because we are free to choose any division ring. In fact, decoupling vectors into $\cat I$ and $\cat V$ allows engineers to mix-and-match index and value categories as desired. Our two restrictions, that $\cat I$ be Cartesian and that $\cat V$ be a division ring, are extremely broad. For instance, binary values are used in BSC, with XOR and AND acting as our multiply and divide, respectively, to give us our division ring.

As a final exercise, we present the full description of a finite-dimensional FHRR using our formalisation.
\begin{example}
Let $\cat I$ be the finite category whose objects are natural numbers, and whose arrows represent the $\leq$ relationship. Let $a \IMon b$ be $\min(a,b)$. Also, let $\cat V = (\C, \cdot, +, 1, 0, |-|_\leq)$. Then $[\cat I, \cat V]$ are complex-valued vectors, with optimal bind and bundle being \[v \otimes w =\text{Ran}_{\IMon} v \overline{ \otimes } w = \int_i v_i \cdot w_i\] and \[v \oplus w =\text{Ran}_{\IMon} v \overline{ \oplus } w = \int_i v_i + w_i\] respectively. Further, we have that \[\simi(v, w) = \int^i v^{\dagger}_i \cdot w_i = \sum_{i \in \ob(\cat I)} \overline{v_i} \cdot w_i.\]
\end{example}

\section{Discussion and Future Work}
\label{sec:disc}

The first thing we want to note is that some of the binding and bundling operations in Table~\ref{tab:comparison} are actually inconsistent with our formalisation. In particular, it's clear that the various modifications to bind and bundle (thresholding, normalisation, etc.) don't arise from division rings. In these cases, such modifications are historically for \emph{modelling}, rather than \emph{computational}, purposes. For example, thresholding is often used to model the kind of \enquote{saturation} that neurons exhibit. This means that engineers add extra mechanics to VSA implementations when the goal is modelling cognitive biases/phenomena. We hope that our formalisation helps engineers design VSAs when computation is their only concern.

Second, the worked examples in Section~\ref{sec:examples} largely validate current VSA models. Example~\ref{ex:pow-set} acknowledges that such an indexing set is well-behaved with any division ring. Finding new possible VSAs using our formalisation is left for future work. Future work can also attempt to generalise the similarity measures we discuss beyond inner products, or characterise why additional generality may not be required for VSAs. Future applications of our theory will also explore the relationship that the similarity function has with bind and bundle.

Another result we wish to work towards is axiomatising a category,~\VSA, in a manner similar to the work on Hilbert spaces in~\cite{heunen2022axioms, heunen2022axiomscon, dimeglio2024dagger}. If possible, we would then be interested to see if it is possible to express an adjunction between these two categories. Doing so would formalise the analogy between VSAs and Hilbert spaces that has long been observed. It may be that there's no meaningful difference between the two, but even if VSAs are just finite-dimensional Hilbert spaces, current axiomatisations of finite-dimensional Hilbert spaces are limited to ones with contractive mappings.

Finally, this work has largely ignored how most of the models built using VSAs are deeply compositional in nature. Appendix~\ref{app:ex} includes instances of compositional data structures for the interested reader (in particular, Appendix~\ref{app:structs}). This clearly indicates that there is a wealth of work that can be done to ground VSA applications in category theory. For instance, modelling \emph{analogy} is a recurring application in the VSA literature~\cite[provides the canonical ``what is the dollar of mexico?'' example]{kanerva2009hyperdimensional}, and there is a clear opportunity to understand this notion more rigorously. Future work will directly explore these kinds of topics more precisely in categorical terms. From an engineering perspective, a significant achievement towards this end would be the construction of a VSA that is purely functional\footnote{\enquote{Functional} in the programming sense, rather than set theoretic.} in nature.

\section{Conclusion}
\label{sec:conc}

In summary, we've proposed the first description of VSAs explicitly using categorical language, to the best of our knowledge. We then generalise from vectors to co-presheaves, and use Kan extensions to guide us our choice of bind and bundle operations. We suspect that other researchers already have the categorical tools to refine our proposed formalisation and invite comments and critiques of this nature. We also provide directions for future work in the hope that this work is just a first step towards a more comprehensive categorical foundation of VSAs.

\section*{Acknowledgements}
\label{sec:tyvm}
NPS would like to thank Clifton Cunningham for first introducing them to categories, and Geoff Cruttwell, Bartosz Milewski, and Priyaa Varshinee Srinivasan for discussions that contributed to this work. PMF thanks Nicole Sandra-Yaffa Dumont for discussion that helped refine ideas presented here.

\newpage
\bibliographystyle{eptcs}
\bibliography{main}
\newpage


\appendix

\section{Literature Survey}
\label{app:lit-rev}
We present a literature survey showing that there is almost no research on category theory applied to VSAs. Using a search string of
\begin{center}
    \code{(\enquote{vector symbolic architecture} OR \enquote{vector symbolic algebra} OR \enquote{hyperdimensional computing})\\ AND\\ \enquote{category theory}}
\end{center}
on Google scholar yielded only twelve results as of 2025-03-06. Of these, one was simply indexing all terms that appeared in IEEE Signal Processing Letters for the year 2014. Another was a job posting for a research position. The remaining ten results were proper academic work that we catalogue below. We list these works roughly in order of increasing relevance.
\begin{enumerate}
    \item A textbook by Maurer that discussed modern connectionism in cognitive science~\cite{maurer2021cognitive}. VSAs were mentioned as one model for neuroarchitectures. Category theory appeared only in a reference title and was completely absent from the non-paywalled parts of the document\footnote{We did our best to discern whether category theory was meaningful to the text by reading a provided abstract for every chapter. There was no mention of the topic therein.}.
    \item An extremely broad survey of mathematics by Raeini that mentions the two topics completely separately~\cite{raeini2024golden}.
    \item A paper by Xue \emph{et al.} on distributed microservices discusses category theory at length, but VSAs only appear as a citation when discussing background work~\cite{xue2021reaching}. The same thing occurs in a paper by Leemhuis and \"{O}z\c{c}ep about analogical proportions and \enquote{betweenness}~\cite{leemhuis2023analogical}.
    \item Three works discussed VSAs at length, but no substantial connection to category theory is made. First, there's a demonstration by Summers-Stay that VSAs can perform deductive inference~\cite{summers2020propositional}. Category theory is mentioned once in the closing paragraph, with the author invoking it to argue that conjunctive normal form (CNF) is a necessary part of their model (we are unsure of the connection, since this is not expanded upon). Second, a work by Widdows and Cohen also discusses inference using VSAs~\cite{widdows2015reasoning}. Last, a work by Gayler and Levy argues against localist representations for analogical mappings in neuronal models~\cite{gayler2009distributed}. This work argues that the distributed representations of VSAs make them better suited to doing analogical reasoning than their localist counterparts. The sole mention of category theory is an aside to justify the representational power of graphs.
    \item One work discusses both in detail, but in separate contexts rather than connecting the two. This is a PhD thesis by Qiu that is primarily concerned with graph embeddings and disentanglement~\cite{qiu2023graph}. Here, VSAs serve as inspiration for methods used to perform graph embedding, while category theory is mentioned in a separate speculative chapter on algorithm transfer. No direct connection is made.
    \item Two works by Widdows \emph{et al.} are quite closely related. The first is a broad overview of how various operations used in quantum mechanics and quantum computing can be applied to AI~\cite{widdows2021quantum}. There, a direct connection is made between VSAs and quantum mechanics, through tensors/binding. A second paper discusses the use of vector representations in more detail~\cite{widdows2021should}. Though category theory is only mentioned briefly in that work, it directly acknowledges how category theory relates quantum models to vector representations of semantics and grammar.
\end{enumerate}


\section{Additional Examples of VSA Algorithms}
\label{app:ex}

Here, we provide a few more examples of how VSAs can be used to perform computations and solve problems. First, we provide a very simple example of how to construct and deconstruct tuples. Then we present two more examples that also highlight reasons why practitioners may wish to use commutative binding versus non-commutative binding.

\subsection{Constructing and Deconstructing Tuples}
\label{sec:tuple}

Sometimes, individuals construct bundles that have more structure than an unordered set. We can, for example, create a tuple with two roles, \texttt{first} and \texttt{second}. To construct that tuple from two vectors, $v_{1}$, $v_{2}$, we would write
\begin{align}
    w = (\mathtt{first} \otimes v_1) \oplus (\mathtt{second} \otimes v_{2}).
\end{align}
where binding takes precedence over bundling for order of operations (we've added parentheses simply for clarity).

Then, to recover any element, we would unbind the corresponding role vector:
\begin{align*}
    v_{1} &\sim \mathtt{first} \oslash w\\
    v_{2} &\sim \mathtt{second} \oslash w
\end{align*}

\subsection{An Argument for Commutative Binding: Representing Numbers}
\label{app:add}

In MAP and (F)HRR, integers and real-valued data can be represented through iterative applications of operations, not unlike using a successor operation to construct the naturals. However, they may be more easily represented by simply binding a given vector iteratively. For example, using an initial vector, $x$, one can represent an integer, $n \in \mathbb{Z}^{+}$,  in the (F)HRR algebras by binding $x$ with itself $n$ times, denoted:
\begin{align}
    \phi[n] &:= x \otimes \ldots \otimes x\\
    &= \overset{n}{\underset{i=1}{\otimes}}x,
\end{align}
where $\phi[n]$ denotes the vector representation of $n$.

For VSAs where vectors are their own inverse (\emph{e.g.}, BSC, MAP), the successor encoding is required. Integers are represented by applying the braiding operation an integer number of times to the initial vector:
\begin{align}
    \phi[n] &:= \braid(\braid(\ldots x \ldots))\\
    &= \braid^{n}x.
\end{align}

In commutative algebras, like the first example, it is the case that $\phi[n+m] = \phi[n]\otimes\phi[m] = \phi[m]\otimes\phi[n] = \phi[m+n]$. In non-commutative algebras this relationship may not hold for all elements. In theory, one could exploit superposition and the pseudo-orthogonality to define a bundle that is the equivalence class of all additions, but this would place constraints on the lower bound on the vector dimension to ensure that bundle capacity is not reached. This problem would be further complicated in the case of \emph{fractional} binding~\cite{plate1992holographic,komer2019neural}, where integer binding is naturally extended to the reals through element-wise exponentiation (computed in the Fourier domain in the case of circular convolution binding), although it is not immediately clear what the appropriate translation of fractional binding would be to the non-commutative algebras.

\subsection{An Argument Against Commutative Binding: Composing Data Structures}
\label{app:structs}

Data structures can be defined in VSAs.  We had previously discussed slot-filler relationships, like the example of constructing a tuple in Section \ref{sec:tuple}, but it is also possible to construct data structures that have more detailed ordering, such as lists or trees. A list of items $(x_{1},\ldots,x_{n})$ can be constructed using bundling and repeated application of a braiding operation, $\rho x:= \braid(x)$ as follows:
\begin{equation}
    \mathrm{list}(x_{1},\ldots,x_{n}) = x_{1} \oplus \rho x_{2} \oplus \rho\rho x_{3} \oplus \ldots \oplus \rho^{n-1}x_{n}
\end{equation}
where $\rho^{k}$ indicates $k \in \mathbb{Z}^{+}$ applications of the braiding operation.  Individual elements can be selected  This construction is made without relying on the binding operation, although such guarding is possible in VSAs with commutative binding operations.  This is accomplished by binding with a \enquote{guard vector}, $h$, that has been iteratively bound with itself:
\begin{equation}
    \mathrm{list}(x_{1},\ldots,x_{n}) = x_{1} \oplus h\otimes x_{2} \oplus \ldots \oplus h^{n-1}\otimes x_{n-1},
\end{equation}
where $h^{k}$ is shorthand for $\underset{i=1}{\overset{k}{\otimes}h}$. Problems with this approach become more apparent when constructing, for example, a binary tree. In this case, one would require two braiding operations, $\rho_{L}$ and $\rho_{R}$, which are typically implemented as two different permutation matrices.  Storing data elements in the leaves of a tree with depth 2 would be implemented:
\begin{equation}
    \mathrm{tree}(x_{1},x_{2},x_{3},x_{4}) = \rho_{L}\rho_{L}x_{1} \oplus \rho_{L}\rho_{R}x_{2} \oplus \rho_{R}\rho_{L}x_{3} \oplus \rho_{R}\rho_{R}x_{4}.
\end{equation}

To construct a tree using guard vectors we would similarity define $h_{L}$ and $h_{R}$, and construct the tree:
\begin{equation}
    \mathrm{tree}(x_{1},x_{2},x_{3},x_{4}) = h_{L}\otimes h_{L}\otimes x_{1} \oplus h_{L}\otimes h_{R}\otimes x_{2} \oplus h_{R}\otimes h_{L}\otimes x_{3} \oplus h_{R} \otimes h_{R}\otimes x_{4}.
\end{equation}
But if $\otimes$ commutes, then $h_{L}\otimes h_{R} = h_{R}\otimes h_{L}$, making it impossible to disambiguate the two middle leaves of the tree. Gayler~\cite{gayler1998multiplicative} proposed a braiding operator for precisely this purpose. Plate suggested that multiple non-commutative, similarity preserving binding operators may exist, noting systematically permuting one of the arguments of a commutative binding operation as one example~\cite[\textsection 3.6.7]{plate2003holographic}. Gosmann and Eliasmith~\cite{gosmann2019vector} propose the VTB as one non-commutative binding operation, simplifying the number of operations required in the VSA to impose structure.


\end{document}